\documentclass[lettersize,journal]{IEEEtran}
\usepackage{amsmath,amsfonts}
\usepackage{algorithmic}
\usepackage{algorithm}
\usepackage{array}
\usepackage[caption=false,font=normalsize,labelfont=sf,textfont=sf]{subfig}
\usepackage{textcomp}
\usepackage{stfloats}
\usepackage{url}
\usepackage{verbatim}
\usepackage{graphicx}
\usepackage{cite}
\usepackage{xcolor}
\usepackage{multirow}
\usepackage{makecell}
\usepackage{booktabs}
\begin{document}
\title{DEPTHOR++: Robust Depth Enhancement from a Real-World Lightweight dToF and RGB Guidance}
%
%
%
\author{\textbf{Jijun Xiang}, \textbf{Longliang Liu}, \textbf{Xuan Zhu}, \textbf{Xianqi Wang}, \textbf{Min Lin}, \textbf{Xin Yang}, ~\IEEEmembership{Member,~IEEE}
\thanks{Jijun Xiang, Longliang Liu, Xuan Zhu, Xianqi Wang, Min Lin, Xin Yang are with the School
of Electronic Information and Communications,
Huazhong University of Science and Technology, Wuhan 430074, China (E-mail:\{jijunx, longliangl, xuanzhu, xianqiw, minlin, xinyang2014\}@hust.edu.cn).}

\thanks{Corresponding author: Xin Yang.}}

\maketitle

\begin{abstract}
Depth enhancement, which converts raw dToF signals into dense depth maps using RGB guidance, is crucial for improving depth perception in high-precision tasks such as 3D reconstruction and SLAM. However, existing methods often assume ideal dToF inputs and perfect dToF-RGB alignment, overlooking calibration errors and anomalies, thus limiting real-world applicability. This work systematically analyzes the noise characteristics of real-world lightweight dToF sensors and proposes a practical and novel depth completion framework—DEPTHOR++, which enhances robustness to noisy dToF inputs from three key aspects. First, we introduce a simulation method based on synthetic datasets to generate realistic training samples for robust model training. Second, we propose a learnable-parameter-free anomaly detection mechanism to identify and remove erroneous dToF measurements, preventing misleading propagation during completion. Third, we design a depth completion network tailored to noisy dToF inputs, which integrates RGB images and pre-trained monocular depth estimation priors to improve depth recovery in challenging regions. On the ZJU-L5 dataset and real-world samples, our training strategy significantly boosts existing depth completion models, with our model achieving state-of-the-art performance, improving RMSE and Rel by 22\% and 11\% on average. On the Mirror3D-NYU dataset, by incorporating the anomaly detection method, our model improves upon the previous SOTA by 37\% in mirror regions. On the Hammer dataset, using simulated low-cost dToF data from RealSense L515, our method surpasses the L515 measurements with an average gain of 22\%, demonstrating its potential to enable low-cost sensors to outperform higher-end devices. Qualitative results across diverse real-world datasets further validate the effectiveness and generalizability of our approach. Code of DEPTHOR is available at: \textcolor{magenta}{https://github.com/ShadowBbBb/Depthor}, and DEPTHOR++ will be released upon the publicity of the paper.

\end{abstract}


\section{Introduction}
\label{sec:intro}

\IEEEPARstart{D}{irect} Time-of-Flight (dToF) sensors are widely deployed on mobile devices for tasks such as autofocus and obstacle detection, owing to their compact size and low power consumption. However, the depth measurements they provide are typically too coarse for high-precision applications like 3D reconstruction~\cite{bundlefusion, Sr-lio, Sr-livo} and SLAM~\cite{dtofslam, kinectfusion, SDV-LOAM}. To address this limitation, depth enhancement has been proposed to reconstruct high-resolution depth maps from raw dToF signals, using accompanying RGB images as guidance.

According to the sensor's data format, depth enhancement methods fall into two categories: depth completion and depth super-resolution. Depth completion methods~\cite{penet,completionformer,bpnet} take high-resolution depth maps with sparsely distributed valid pixels as inputs and then reconstruct dense depth maps by propagating these points using geometric reasoning and RGB guidance. In comparison, depth super-resolution methods~\cite{xia2020generating,promptda} operate on low-resolution dense depth maps where each depth value corresponds to a local image region, and upsample the depth map to match RGB resolution through cross-modal guidance.

Existing dToF depth enhancement methods~\cite{deltar,dvsr,cfpnet} are typically designed for depth super-resolution and rely on two key assumptions: \emph{(1) ideal calibration between the camera and sensor, providing precise RGB-dToF correspondence} and \emph{(2) reliable operation of the dToF sensor, returning accurate values}. However, these assumptions often fail in real-world scenarios. Through a systematic analysis of RGB-dToF samples collected from a mobile device, we observed that calibration errors are inevitable and may accumulate over time, leading to nontrivial spatial misalignments. Furthermore, due to the sensor's imaging principles, certain regions suffer from anomalies such as signal loss or error values.

We argue that for enhancing depth from a real-world, lightweight dToF sensor, it is more appropriate to project the raw dToF signals into high-resolution sparse depth maps using device parameters and formulate the task as a sparse depth completion problem. The key motivation behind this argument is that in this formulation, both calibration errors and signal anomalies appear as depth point inconsistencies at global or local scales. This allows us to focus solely on improving the robustness of the depth completion model against anomalies, eliminating the need for complex region correspondences and thereby relaxing the restrictive assumptions commonly made in prior works. For depth completion with noisy input, the problem centers around two key challenges: \textit{(1) how to robustly train the model in the absence of accurate depth ground truth;} and \textit{(2) how to reliably complete the depth map when the input dToF data contains substantial noise.}


In terms of training, existing methods~\cite{tpvd,deltar} and datasets~\cite{nyuv2,scannet,arkitscenes,kitti} typically use low-cost sensor data as input and high-cost sensor data as ground truth for supervision, focusing on improving the output of low-cost sensors to approximate the performance of high-end sensors. This setup introduces two key challenges: First, high-precision sensors often share similar anomalies with low-cost ones, limiting the model’s ability to learn effective corrections and thus capping its performance at the high-precision sensor's level. Although some datasets~\cite{hammer, scrream} provide high-quality ground truth using industrial-grade scanners and auxiliary techniques, the complexity and high cost of data acquisition significantly limit their diversity and scale, making them more suitable for evaluation rather than training. Second, real datasets are usually tailored to specific types of sensors, limiting their generalizability and reusability across other sensor platforms, while collecting such data for a new sensor is costly.

\begin{figure*}[!t]
	\centering
	\includegraphics[width=1\linewidth]{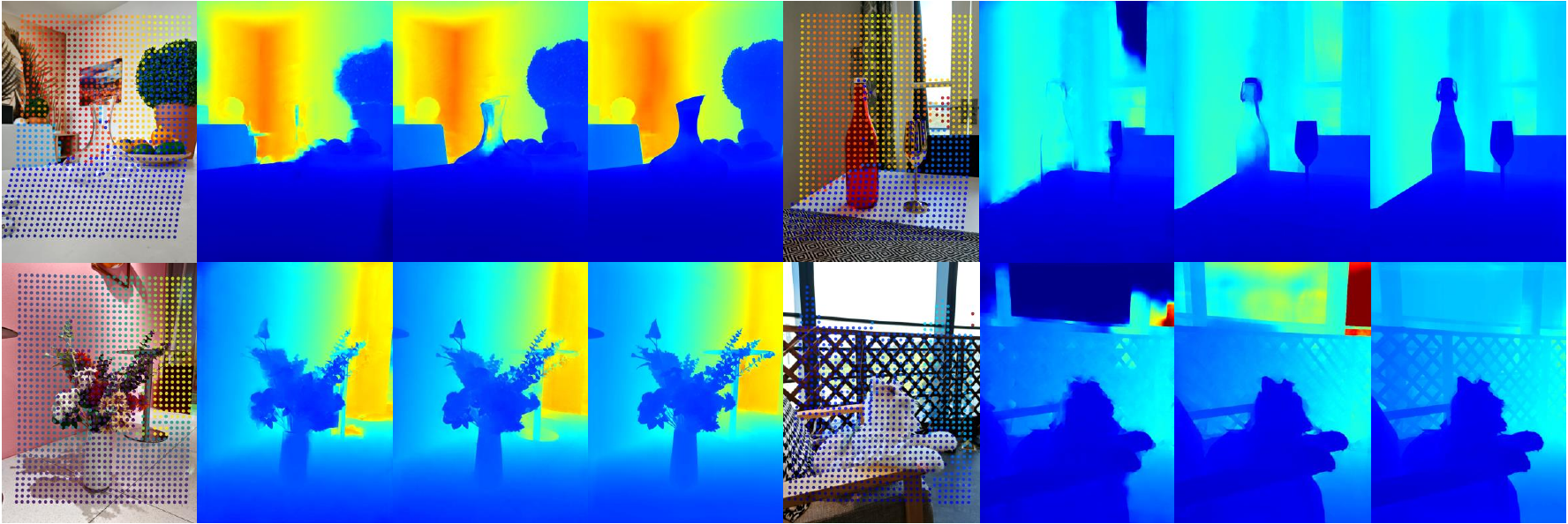}
	\caption{\textbf{Effect of our training strategy and our depth completion model (without anomaly detection).} From left to right are: RGB-dToF, predictions of a lightweight PENet~\cite{penet}, the same PENet with our training strategy, and our model with our training strategy. Our training strategy improves the performance of existing methods on real-world data. Our model further enhances predictions in challenging regions.}
	\label{fig:implicit}
\end{figure*}

To address the above challenges, we propose to train on synthetic datasets~\cite{hypersim,tartanair,dynamicreplica}, which offer accurate and detailed ground truth for supervised learning. Based on a systematic analysis of real-world RGB-dToF samples, we introduce a simulation method to further align the synthetic data with real-world dToF characteristics. This method accounts for four key aspects: global distribution, abnormal regions, calibration errors, and random noise, which can be easily adapted to different dToF sensors.

From the perspective of depth completion, existing methods rarely consider the unique depth pattern of dToF and typically assume ideal sensor behavior, leading to limited performance when facing the real-world dToF data with uniform distribution, misaligned with the RGB image, limited field-of-view, and anomalies in specular, transparent, and low-reflectivity regions. Even with our proposed training strategy, the simulated sensor noise cannot fully capture the diversity of real dToF noise patterns, making it necessary to design dedicated methods to further enhance robustness against noisy dToF inputs.

To address this, we first design an anomaly detection mechanism that explicitly identifies and masks erroneous measurements before depth completion. This is motivated by our observation that while both missing and erroneous signals are theoretically considered anomalies, in practice, missing points are significantly easier to handle. Therefore, explicitly detecting errors and converting them into missing values can further improve the model’s robustness by avoiding the misleading propagation of erroneous measurements.

Previous studies~\cite{mirror3d, cleargrasp, tdcnet} have explored anomaly detection for dense, high-resolution depth maps. However, these methods often target specific scenarios such as mirrors or transparent surfaces, making them task-specific and difficult to generalize across applications. Moreover, operations on dense depth maps typically incur high computational cost.

In this paper, we introduce a general-purpose anomaly detection method tailored for sparse depth points. We observe that the human visual system detects depth anomalies by leveraging relative depth perception and regional consistency. Recent monocular depth estimation (MDE) models exhibit similar capabilities in several aspects:
(a) strong predictive performance in challenging regions enabled by large-scale, high-quality training data;
(b) highly accurate relative depth relationships, despite some imprecision in metric values;
(c) better boundary preservation compared to most depth sensors.

Inspired by these insights, we employ the MDE model to approximate the human visual system and propose a zero-learnable parameter anomaly detection method. Specifically, we compute two types of anomaly scores for each sparse depth point:
(1) whether its global rank in the sensor measurements is consistent with its rank in the MDE-based relative depth map;
(2) whether its local depth relationships with neighboring points remain consistent between the sensor and MDE.
We then combine the Spearman rank correlation coefficient and the Otsu thresholding to dynamically generate the final anomaly mask. As the MDE module is already embedded in our system and its outputs are retained, this anomaly detection process introduces only minimal computational overhead.

In addition, we design a simple yet efficient depth completion network capable of handling various types of sensor input noise. A pre-trained monocular depth estimation (MDE) model is introduced to provide relative depth relationships and contextual information, enhancing prediction accuracy in challenging regions. The network consists of two stages: (1) Multimodal fusion stage: An encoder-decoder architecture is employed to extract and fuse RGB and depth features, generating an initial depth prediction. (2) Refinement stage: We fuse features from the MDE model and the decoder to construct a mixed affinity map, which is then used to refine the initial prediction. This architecture maintains computational efficiency while significantly improving the model’s adaptability to complex noise patterns and overall depth prediction accuracy.

\begin{figure*}[!t]
	\centering
	\includegraphics[width=1\linewidth]{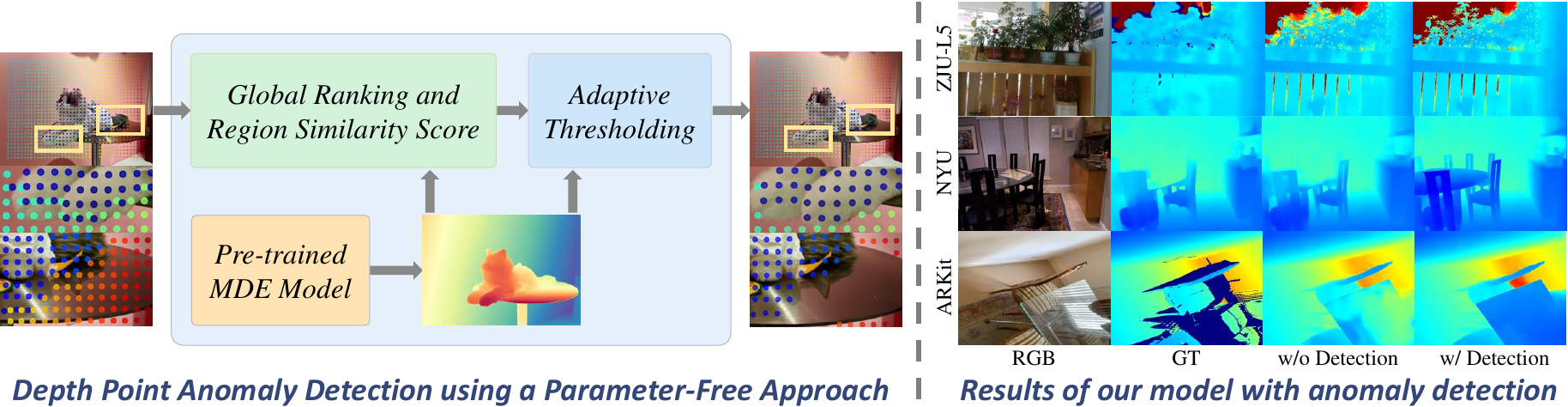}
	\caption{Left: Overview of the anomaly detection method. Right: Results of combining the detection method with our depth completion model, the input dToF points are sampled from ground truth collected by high-precision sensors.}
	\label{fig:explicit}
\end{figure*}

We conducted experiments on multiple datasets to validate the effectiveness of our approach: First, on the ZJU-L5 dataset and a more challenging set of dToF samples, our training strategy significantly enhances existing depth completion models, achieving results comparable to depth super-resolution methods. Furthermore, our model outperforms all types of state-of-the-art methods, improving Rel and RMSE by 27\%, 18\% and 15\%, 7\%, respectively. Second, on the mirror regions of the Mirror3D-NYU dataset, by incorporating the anomaly detection method, our model outperforms the previous SOTA by 32\% and 42\% for RMSE and Rel. Third, on the Hammer dataset, we simulated dToF signals by sampling 30$\times$40 points from depth maps generated by RealSense L515. Our full method outperforms the original L515 measurements, improving RMSE and Rel by 31\% and 14\%, demonstrating its potential to enable low-cost sensors to outperform higher-end devices. Qualitative results on additional real-world datasets, including NYUv2 and ARKitScenes, further confirmed the generalizability of our approach.

In summary, our main contributions are as follows:
\begin{itemize}
	\item We propose DEPTHOR++, a comprehensive solution for real-world dToF enhancement, comprising both implicit learning (training strategy, depth completion model) and explicit detection (depth point anomaly detection).
	
	\item We propose a noise-robust training strategy based on a novel dToF simulation method on synthetic datasets.
	
	\item We design a depth completion model that revises components in existing architectures to better suit the characteristics of dToF and real-world anomalies. Additionally, it integrates MDE prior at multiple stages to enhance predictions in challenging regions.
	
	\item We introduce a depth point anomaly detection method by leveraging the ranking consistency and regional similarity between dToF and MDE, which explicitly detects and masks errors to further enhance robustness.
\end{itemize}

\section{Related Work}
\label{sec:related}

\subsection{Depth Sensor Dataset}
\label{sec:related_dataset}
Early datasets~\cite{nyuv2,kitti} laid the foundation for depth enhancement. For instance, NYU Depth v2~\cite{nyuv2} uses Microsoft Kinect v1 to collect ground truth with standard test inputs consisting of 500 randomly sampled points. However, due to the perfect alignment between inputs and GT, metrics on these datasets fail to adequately reflect a model’s ability to handle potential real-world anomalies.

Subsequent methods have increasingly emphasized the enhancement of low-cost sensors by simultaneously equipping both low- and high-precision devices. This setup mitigates the issue of perfect alignment between input and ground truth. Consequently, the evaluation metrics reflect how well a depth enhancement model can enhance a low-cost sensor to approach high-precision ones. For instance, ZJU-L5~\cite{deltar} uses ST VL53L5CX and Intel RealSense 435i to acquire raw inputs and ground truth, while TOFDC~\cite{tpvd} employs a Huawei P30 Pro and Helios ToF camera. Furthermore, industrial-grade sensors have also been widely employed to acquire more accurate ground truth, such as iPad Pro and Faro Focus S70 used in ARKitScenes~\cite{arkitscenes}. However, even the most precise sensors may still produce anomalies, particularly in boundary regions or scenes with complex materials.

Recent datasets~\cite{hammer,scrream} have attempted to obtain reliable and real ground truth. For instance, the Hammer dataset~\cite{hammer} collects raw depth data from Intel RealSense L515 (dToF), Lucid Helios HLS003S-001 (iToF), and Intel RealSense D435 (active stereo), all of which contain inherent anomalies. The dataset employs Einscan-SP and Artec Eva scanners to obtain high-precision ground truth, supplemented by 3D rendering, AESUB Blue vanishing spray, and temporarily attached small markers. Due to the complexity of these acquisition techniques, such datasets often exhibit limited diversity and coverage, making them more suitable for evaluation.

\subsection{Depth Completion} 
\label{sec:related_completion}
Conventional depth completion methods can be categorized into encoder-decoder~\cite{acmnet,completionformer,penet} and affinity propagation approaches~\cite{lrru,cspn++,cspn}.
Recent approaches have made significant progress in addressing practicality through various techniques, including 3d operation~\cite{bpnet, tpvd}, test-time adaptation~\cite{testtime}, specialized modules~\cite{spnet, mspn}, novel architectures~\cite{omnidc}, and integration with MDE models~\cite{steeredmarigold, marigolddc}.

However, assessing the applicability of existing methods to real-world dToF data remains a challenge. First, existing methods are typically trained and evaluated on real-world datasets. However, the sensors used to collect ground truth often exhibit similar anomaly patterns to dToF. As a result, models struggle to learn how to correct such anomalies, and evaluation metrics fail to effectively measure improvements. Second, the unique characteristics of dToF are rarely considered. To the best of our knowledge, only methods from the MIPI competition~\cite{mipi2022,mipi2023,emdc,svdc} have attempted to simulate the uniform distribution using grid sampling. Therefore, many designs that are considered effective in other depth modalities are not suitable for dToF.

\subsection{Depth Super-resolution}
\label{sec:related_resolution}
For dToF depth super-resolution, Deltar~\cite{deltar} proposed a dual-branch network that utilizes PointNet to extract dToF features and employs a transformer-based fusion module to integrate RGB and depth information. Building upon this, CFPNet~\cite{cfpnet} addressed the limited FoV coverage of dToF sensors by incorporating large convolution kernels and cross-attention mechanisms to enhance predictions in border regions. DVSR~\cite{dvsr} specifically addresses video sequences, using optical flow and deformable convolutions to aggregate multi-frame information, thereby enhancing prediction consistency. 
The dToF simulation in these approaches usually begins by computing the depth histogram within a given region of the ground truth, followed by computing the target signal of dToF (e.g., mean, peak, variance, rebin histograms). Among these, DVSR addresses the issue of signal loss in low-reflectivity regions by approximating the diffuse reflectance using the mean value of the corresponding RGB patch.

It is noteworthy that existing methods typically rely on accurate RGB-dToF correspondence. For example, the ZJU-L5 dataset provides the coordinates of the RGB region corresponding to each dToF signal, and both Deltar and CFPNet leverage these coordinates to guide their feature aggregation modules. DVSR assumes that the dToF data is uniformly distributed, dividing a $480\times640$ image into $30\times40$ patches and directly simulating the dToF from depth GT corresponding to each patch. When real-world devices fail to provide accurate correspondences, the performance of these methods deteriorates significantly.

\subsection{Monocular Depth Estimation}
\label{sec:related_mde}
Early methods trained on a single dataset predict metric depth, but due to the inherent lack of depth scale information, these methods have poor generalization across different datasets. Midas~\cite{midas} introduced an affine-invariant loss to predict the inverse depth, making the model focus on relative relationships rather than absolute values, thus mitigating the impact of scale shifts between different datasets.

Recent models~\cite{depthanythingv1,depthanythingv2,metric3d,metric3dv2,marigold} have significantly advanced this field with methods like pseudo-label generation, diffusion model priors, and additional normal supervision. Among these, Marigold~\cite{marigold} achieves high detail in depth map outputs; however, its reliance on a diffusion model results in long inference times, which conflicts with many dToF application scenarios. Metric3d~\cite{metric3dv2}, by introducing a normalized camera model and normal supervision, enables the model to output scaled depth with generalization. In contrast, Depth Anything series~\cite{depthanythingv1,depthanythingv2} outputs inverse depth. Through methods like pseudo-label generation and distillation learning, it provides state-of-the-art results with fast inference speed, making it a compatible choice for many depth-related downstream tasks. 

\subsection{Depth Anomaly Detection}
\label{sec:related_filter}
Most existing methods are task-specific for mirrors or transparent surfaces. Instead of directly identifying anomalies in the depth domain, these approaches typically rely on RGB images to detect specific categories, using this semantic guidance to refine depth maps. For instance, Mirror3D~\cite{mirror3d} proposes a model that jointly predicts mirror masks and surface normals from RGB images, followed by propagating valid boundary depth values across mirror regions. The authors further annotate mirror regions in existing real-world datasets and introduce the Mirror3D dataset for benchmarking. Similarly, ClearGrasp~\cite{cleargrasp} addresses transparent object grasping by leveraging multiple models to estimate surface normals, object boundaries, and transparency masks from RGB images, which are then fused to refine depth maps. More recent approach TDCNet~\cite{tdcnet} has simplified this pipeline by designing end-to-end networks that take RGB images and dense depth maps as input to directly output refined depth predictions.

However, these methods are inherently limited by task-specific designs, restricting their generalizability to other anomaly types, and their operation over dense, high-resolution depth maps further increases computational complexity, also making them unsuitable for sparse depth points.

\section{Method}
\label{sec:method}
In this section, we detail the method of DEPTHOR++, which consists of preliminary of dToF imaging in Section~\ref{sec:method_preliminary}, training strategy with dToF simulation in Section~\ref{sec:method_strategy}, depth point anomaly detection in Section~\ref{sec:method_filter}, depth completion model integrating MDE in Section~\ref{sec:method_model} and implementation details in Section~\ref{sec:method_detail}.

\subsection{Preliminary: dToF imaging}
\label{sec:method_preliminary}
\begin{figure}[htbp]
	\centering
	\includegraphics[width=0.9\linewidth]{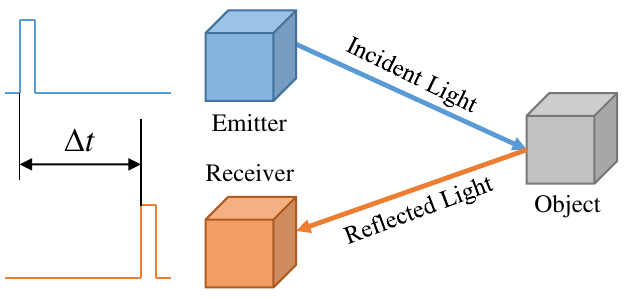}
	\caption{Imaging principle of direct Time-of-Flight sensor}
	\label{fig:principle}
\end{figure}

As shown in Figure~\ref{fig:principle}, a pulsed laser generates a short light pulse and emits it into the scene. The pulse scatters, and some photons are reflected back to the dToF detector. The depth is then determined by the formula $d=\Delta t\cdot c/2$, where $\Delta t$ is the time difference between laser emission and reception, and $c$ is the speed of light. Each dToF pixel captures all scene points reflected within its individual field-of-view (iFoV) using time-correlated single-photon counting. The iFoV is determined by the sensor’s total field-of-view (FoV) and spatial resolution, returning the peak signal detected within that range. More details please refer to~\cite{dvsr, dtof0, dtof1}.

\subsection{Training Strategy with dToF Simulation}
\label{sec:method_strategy}
We collected a set of RGB-dToF samples using an Honor Magic6 Ultra to analyze real-world dToF data. Using the intrinsic and extrinsic parameters of dToF and camera, along with calibration matrices, we project dToF signals into a high-resolution sparse depth map, and Figure~\ref{fig:abnormal} (a) shows an ideal sample. In Figure~\ref{fig:abnormal} (b) - (e), we display exemplar samples which exhibit typical distribution characteristics and potential anomalies of dToF sensors, which are also described as follows:

\begin{figure}[htbp]
	\centering
	\includegraphics[width=1\linewidth]{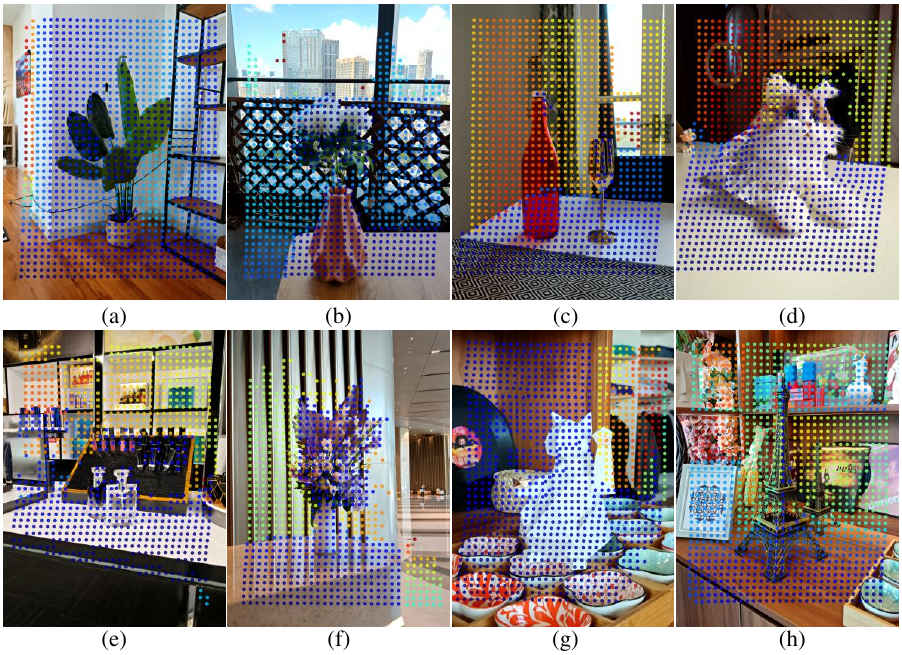}
	\caption{Ideal and anomalous real-world RGB-dToF samples we collected.}
	\label{fig:abnormal}
\end{figure}

\noindent{\textbf{Overall Distribution.}} Due to the FoV difference, the captured depth does not cover the whole image. Moreover, the projected depth points are uniformly distributed and inherently imprecise, as they theoretically present peak values within a defined iFoV, as shown in Figure~\ref{fig:abnormal} (h).

\noindent{\textbf{Abnormal Regions.}} Limited to imaging principles, dToF is prone to anomalies in the following regions:

\begin{enumerate}
	\item \textit{Non-Lambertian}: In the specular area (Figure~\ref{fig:abnormal} (g)), multi-path effects will introduce error measurements. In transparent surfaces, photons may pass through, leading to signal loss (Figure~\ref{fig:abnormal} (b)) or returning further values (Figure~\ref{fig:abnormal} (c)). 
	\item \textit{Low-reflectivity}: In low-light conditions or black surfaces (Figure~\ref{fig:abnormal} (e)), photons are likely to be absorbed rather than reflected, leading to signal loss.
	\item \textit{Long-distance}: Photons are more susceptible to noise at greater distances, and may be lost entirely if they exceed the maximum reception time (Figure~\ref{fig:abnormal} (f)).
\end{enumerate}

\noindent{\textbf{Calibration Errors.}} It manifests as regional shifts after 
projection. Empirically, we observed that foreground points generally project with high precision, while background points often experience a noticeable shift (Figure~\ref{fig:abnormal} (d)). 

The ideal training setup would use the above samples as input, along with accurate ground truth for supervision and evaluation. However, acquiring high-quality GT introduces nontrivial overhead and is difficult in real-world scenarios. In this work, we propose to simulate the dToF data from the GT of synthetic datasets as a substitute.

We design our simulation method based on the aforementioned dToF characteristics: For overall distribution, we perform random translations and rotations on the depth GT within the iFoV range, followed by approximately uniform sampling within the roughly defined FoV. For anomalies in special regions, we categorize them into two types: absence and error. We generate irregularly shaped masks and randomly assign each region an anomaly type to achieve better diversity and uncertainty. For calibration error, we select a percentile from GT as the threshold, treat points above it as background, and apply a random shift (within 0–2 dToF pixels). Additionally, to enhance the model's robustness to random anomalies, we introduced approximately 5\% noise points and 5\% blank points. The depth values of the noise points were randomly assigned within the theoretical detection range.

\begin{figure}[htbp]
	\centering
	\includegraphics[width=1\linewidth]{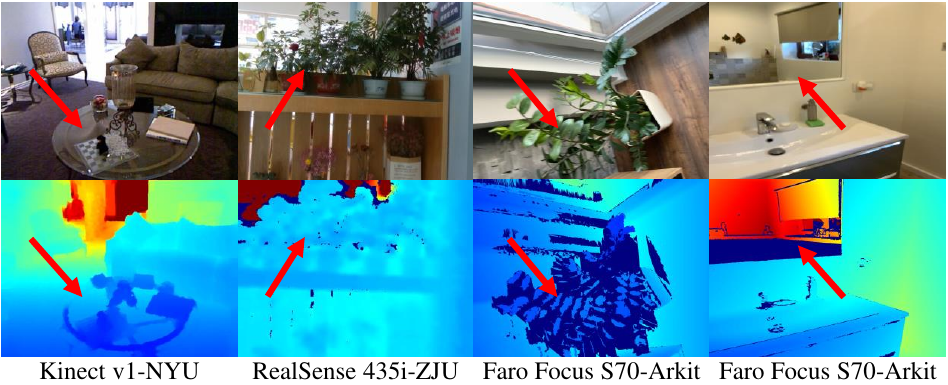}
	\caption{RGB and depth GT of existing real-world datasets. The red arrows indicate unreliable measurements in challenging regions.}
	\label{fig:realdataset}
\end{figure}

Theoretically, our simulation method can also be applied to real-world datasets. However, as shown in  Figure~\ref{fig:realdataset}, we observe that high-precision sensors struggle to produce reliable measurements in the challenging regions we targeted, without the aid of auxiliary techniques, even the most precise Faro Focus S70 scanner used in the ARKitScenes dataset. Therefore, we only simulate dToF data on synthetic datasets and supervise with accurate ground truth, which encourages the model to learn robust features under imperfect data conditions, thereby better adapting to real-world conditions and partially mitigating issues inherent to dToF imaging, as shown in Figure~\ref{fig:implicit}.

\subsection{Depth Point Anomaly Detection}
\label{sec:method_filter}

We consider a sparse set of $N$ depth points $\mathcal{X} = \{x_i = (d_i, r_i, p_i)\}_{i=1}^N$, where each $x_i$ is defined by the absolute depth $d_i$ from the sensor, the relative depth $r_i$ predicted by the MDE model, and the normalized image coordinate $p_i$.

\noindent\textbf{Global Ranking.}  
For each depth point $x_i$, we compute its global rankings in the absolute and relative depth as:
\begin{align}
	G^{\text{abs}}_i &= \frac{1}{N} \sum_{j=1}^{N} \operatorname{sgn}(d_i - d_j), \\
	G^{\text{rel}}_i &= \frac{1}{N} \sum_{j=1}^{N} \operatorname{sgn}(r_i - r_j),
\end{align}
where $\operatorname{sgn}(\cdot)$ denotes the sign function. The ranking inconsistency for $x_i$ is then computed as:
\begin{equation}
	G_i = \tanh\left( \frac{| G^{\text{abs}}_i - G^{\text{rel}}_i |}{\delta} \right),
\end{equation}
where $\tanh(\cdot)$ suppresses minor fluctuations, $\delta=0.5$ is a smoothing parameter.

\noindent\textbf{Region Similarity.}  
For any pair of points $(x_i, x_j)$, we define the scale-invariant depth differences and spatial proximity as:
\begin{align}
	v_{ij}^{\text{abs}} &= \frac{|d_i - d_j|}{d_i + d_j + \varepsilon}, \\
	v_{ij}^{\text{rel}} &= \frac{|r_i - r_j|}{r_i + r_j + \varepsilon}, \\
	w_{ij} &= \exp(-\alpha \|p_i - p_j\|_2),
\end{align}
where $\varepsilon = 10^{-6}$ avoids division by zero, and $\alpha=15$ is a spatial decay constant. We further define the inconsistency between the two depths as:
\begin{equation}
	s_{ij} = w_{ij} \cdot |v_{ij}^{\text{abs}} - v_{ij}^{\text{rel}}|.
\end{equation}

Then, for each point $x_i$, the region-based inconsistency is:
\begin{equation}
	S_i = \frac{1}{N} \sum_{j=1}^{N} s_{ij}.
\end{equation}

\noindent\textbf{Anomaly Score and Adaptive Thresholding.}  
The final anomaly score for point $x_i$ is computed by:
\begin{equation}
	A_i = S_i + G_i.
\end{equation}

We apply the Otsu's method~\cite{otsuthreshold} to compute a threshold $t_{\text{otsu}} = \mathcal{T}_{\text{Otsu}}(A)$. However, as a method originally designed for foreground-background segmentation, the Otsu algorithm assumes a bimodal distribution and tends to produce false positives when anomalies are rare. Therefore, we compute the Spearman rank correlation coefficient $\gamma$~\cite{spearman} between $\{d_i\}$ and $\{r_i\}$ to assess the overall reliability of the depth points. A value of $\rho$ close to 1 indicates high consistency, suggesting the depth measurements are reliable.

Additionally, we define a statistical threshold based on a fixed percentile $p$:
\begin{equation}
	t_{\text{stat}} = \text{TopK}(A, \lfloor p \times N \rfloor).
\end{equation}

The final threshold is defined as a piecewise function of $\gamma$:
\begin{equation}
	t =
	\begin{cases}
		+\infty, & \gamma > 0.95, \\
		w \cdot t_{\text{stat}} + (1 - w) \cdot t_{\text{otsu}}, & 0.85 < \gamma \leq 0.95, \\
		t_{\text{otsu}}, & \gamma \leq 0.85,
	\end{cases}
\end{equation}
where $w = \operatorname{sigmoid}(k(\gamma - u))$ is an interpolation weight ranging from 0 to 1, and $p, k, u$ are hyperparameters. The subdomains are empirically decided. Points with scores larger than $t$ are considered as anomalies.

\subsection{Depth Completion Model Integrating MDE}
\label{sec:method_model}
\begin{figure*}[htbp]
	\centering
	\includegraphics[width=0.9\textwidth]{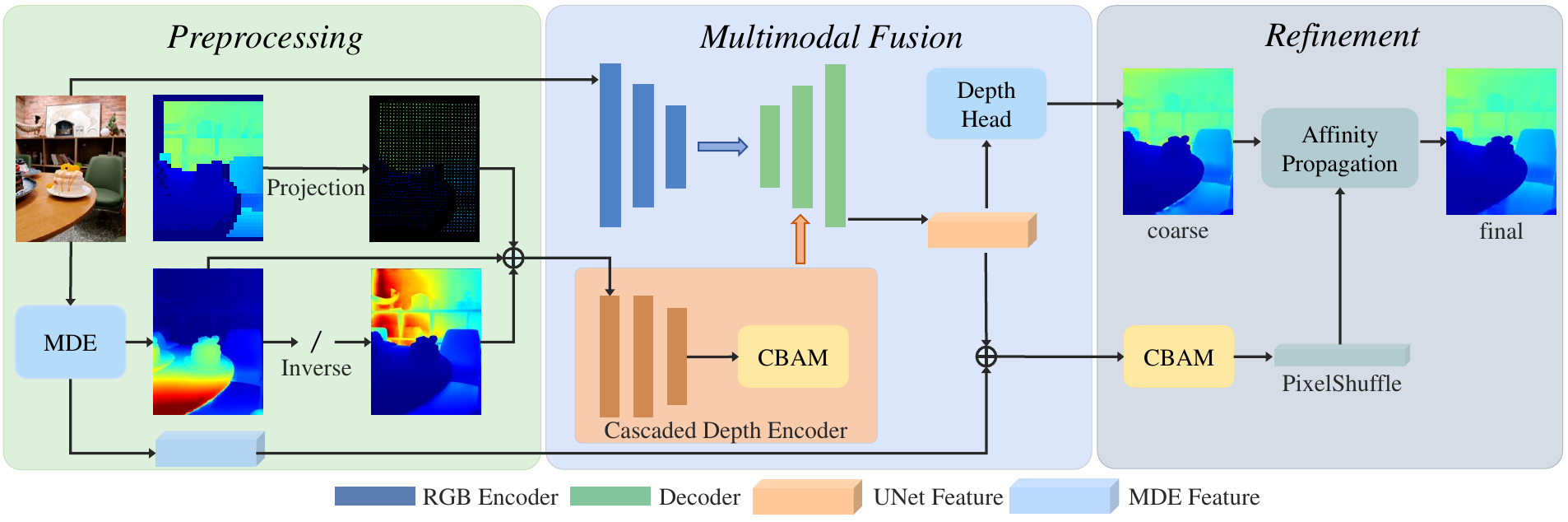}
	\caption{\textbf{Overview of our depth completion model:} We first project dToF signals into sparse depth points, and use a pre-trained MDE model to generate inverse and relative depth maps. In multimodal fusion, we employ a simple encoder-decoder structure to obtain a coarse estimation. In refinement, we update the initial depth map using mixed affinity propagation.}
	\label{fig:structure}
	\vspace{-1em}
\end{figure*}

We formulate the problem as: given a projected sparse depth map $S$, the corresponding RGB image $I$, and the inverse depth map ${D}_{inv}$ and features ${F}_{mde}$ output by the MDE model (based on $I$), the goal is to predict a dense depth map $D$. Figure~\ref{fig:structure} shows the overall structure. 

\noindent \textbf{Multimodal Fusion.} We implement an encoder-decoder network. The encoder extracts multi-resolution features from image and depth separately, which are subsequently fused in the decoder. The fused feature ${F}_{unet}$ is then passed through a depth head to produce an initial depth estimation.

We employed the network from BPNet~\cite{bpnet} as the RGB encoder, progressively downsampling the RGB image and generating feature maps $F_{img}$ at resolutions ranging from 1/2 to 1/32. We modified its architecture and feature dimensions to reduce computational cost and parameters.

For the depth encoder, the inputs consist of \(\{D_{inv}, D_{rel}=1/{{D}_{inv}}, S\}\). We introduce the relative depth map ${D}_{rel}$ to emphasize structural details in distant regions, since they are numerically compressed toward zero in ${D}_{inv}$. Additionally, ${D}_{rel}$ and ${D}_{inv}$ are normalized, $S$ remains unnormalized to retain absolute scale information. With these simple designs, the depth encoder maintains a balance between near and far regions, as well as between relative and absolute depth. To effectively extract depth features $F_{dep}$, we first apply a combination of convolution layers, including large-kernel dilated convolution to enhance the perception of scale information in $S$ and small-kernel downsampling convolution to capture high-frequency details in $D_{rel}$ and $D_{inv}$. Then, we feed the output feature into a CBAM module~\cite{cbam}, where spatial and channel attention are employed for feature enhancement.

In the decoder, we progressively fuse RGB and depth features through convolution and upsampling layers, ultimately producing the decoder feature $F_{unet}$.

Compared to directly regressing a value from features commonly used in depth completion, we introduce the depth head from depth super-resolution~\cite{deltar,adabins}, which allows predefining a fixed depth range (e.g., 0 – 10m) and encourages the model to explore values beyond the observed measurements. Specifically, the depth head uses $F_{unet}$ to generate a set of $N$ non-uniformly normalized depth bins $b$ for each image, along with weighting coefficients $k_i$ for each pixel corresponding to $b_i$. After restoring the depth bins to metric depth using hyperparameters and computing each bin's center $c_i$, the initial depth is computed using the following formula:
\begin{align}
	d=\sum\limits_{i=1}^{N}{{{k}_{i}}{{c}_{i}}} \label{eq:depthhead}
\end{align}

To balance computational cost and accuracy, we set $N$ to 128 and predict the initial depth map at half resolution.

\noindent \textbf{Refinement.} We deploy an affinity propagation module based on CSPN++~\cite{cspn++}, to further refine the initial depth map, such as artifacts in regions without depth point coverage and residual erroneous signals. Unlike previous methods that compute affinity using single-modality features from the decoder, we jointly compute affinity, since the rich semantic information in ${F}_{mde}$ helps correct errors in ${F}_{unet}$ caused by inaccurate depth signals. Meanwhile, incorporating ${F}_{unet}$ mitigates the resolution discrepancies introduced by the Transformer architecture and the lack of scale information in ${F}_{mde}$.

We first interpolate ${F}_{mde}$ to align with the resolution of ${F}_{unet}$. Then, both features are concatenated into a CBAM module and a PixelShuffle \cite{pixelshuffle} layer to upsample to the full resolution. Using this merged feature ${F}_{cspn}$, we calculate mixed affinity ${\omega}_k$ at fixed kernel size $[3,5,7]$. During the propagation, the update process of pixel $i$ under the affinity kernel $k$ at the $t$-th iteration is formulated in \eqref{eq:propagation}. 
\begin{align}
	\hat{D}_{i,k,t} = \omega_{i,k} \hat{D}_{i,t-1} + \sum_{j \in \mathbb{N}_k(i)} \omega_{j,k} \hat{D}_{j,t-1} \label{eq:propagation}
\end{align}

Following BPNet~\cite{bpnet}, we aggregate the outputs across different iterations and affinity kernels using two normalized weights produced by a convolution and softmax layer, as described in \eqref{eq:aggregation}, where $t\in \{0,T/2,T\}$.
\begin{align}
	D = \sum_{t \in T} \tau_t \sum_{k \in \mathcal{K}} \sigma_k \hat{D}_{k,t} \label{eq:aggregation}
\end{align}

Conventional settings typically employ residual connections to add the initial sparse depth map to the coarse prediction before the update. During the iterative update process, point embedding is also often used to directly assign the original sparse depth values to the updated depth map at each iteration. However, since dToF points are not entirely accurate, we remove these settings.

\noindent \textbf{Loss Function.} Following~\cite{deltar}, we employ a scaled affine-invariant loss for supervision, with the expression as follows:
\begin{align}
	L=\alpha \sqrt{\frac{1}{T}\sum\limits_{i}{g_{i}^{2}-}\frac{\lambda }{{{T}^{2}}}{{(\sum\limits_{i}{{{g}_{i}}})}^{2}}} \label{eq:loss}
\end{align}

Where ${{g}_{i}}=\log {{\tilde{d}}_{i}}-\log {{d}_{i}}$, ${{\tilde{d}}_{i}}$, ${{d}_{i}}$, represent the predicted values and ground truth for valid pixel points, respectively, and in all experiments, $\alpha$=10, $\lambda$=0.85. We calculate the loss only for pixels within the sensor's theoretical detection range. 

\subsection{Implementation Details}
\label{sec:method_detail}
To balance inference efficiency and overall performance, we adopt the relatively small version of Depth Anything V2 as the pretrained MDE model, unless otherwise stated in the ablation studies. The MDE model is frozen during training to preserve its generalization ability. We implement our model in PyTorch~\cite{paszke2019pytorch} and train it on 4 Nvidia RTX 3090 GPUs. We adopt AdamW~\cite{adamw} with 0.1 weight decay as the optimizer, and clip the gradient whose ${{l}^{2}}$-norm is larger than 0.1. Our model is trained from scratch in roughly 230K iterations using the OneCycle~\cite{onecycle} learning rate policy, setting the initial learning rate to 1/25 of the maximum learning rate and gradually reducing to 1/100 in the later stages of training. We set the batch size as 12 and the largest learning rate as 0.0003.

Since the anomaly detection method is parameter-free and primarily designed for large-scale erroneous measurements, it is not integrated into the training process. This setting avoids conflicts with batched training and enables the model to process noisy inputs independently. Regarding hyperparameters, we perform a grid search over uniformly sampled candidates and select the best-performing configuration. The final values are set to $p=0.04$, $k=40$, and $u=0.9$.

\section{Experiments}
\label{sec:experiment}
In this section, we validate our method through comprehensive experiments. We began by a introduction of used datasets and metrics in Section~\ref{sec:exp_datasetmetric}, then separately validate the effectiveness of our training strategy (Section~\ref{sec:exp_strategy}), depth completion model (Section~\ref{sec:exp_model}), anomaly detection (Section~\ref{sec:exp_filter}), while ablation studies are presented in Section~\ref{sec:exp_ablation}.

\subsection{Datasets \& Evaluation Metrics}
\label{sec:exp_datasetmetric}
\noindent{\textbf{Hypersim dataset for training.}} We trained our model on the Hypersim~\cite{hypersim} dataset, with 59,544 frames for training and 7,386 frames for testing. For the following three different dToF sensors (testing datasets), we modified the simulation method to ensure a similar distribution. Please refer to the supplementary material for more details.

\noindent{\textbf{ZJU-L5 dataset for testing.}} Deltar~\cite{deltar} employs the ST VL53L5CX (L5) and the Intel RealSense 435i to capture raw dToF data and ground truth, with resolutions of $8\times8$ and $480\times640$. We utilize the provided iFoV to convert each dToF signal into a depth point at the center of its corresponding region, without using the variance information of dToF.

\begin{figure}[htbp]
	\centering
	\includegraphics[width=\columnwidth]{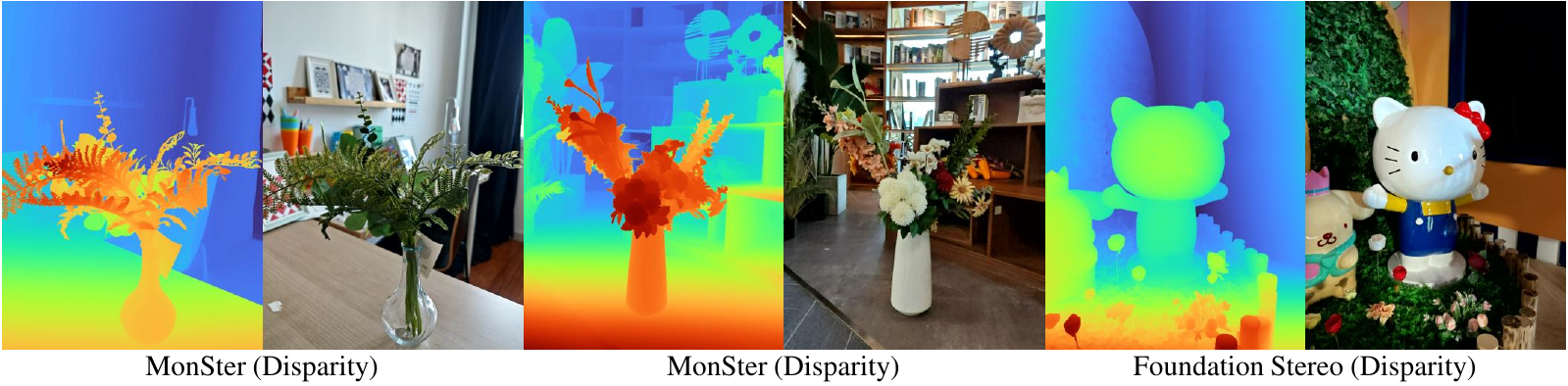}
	\caption{Ground truth from two SOTA stereo methods for our samples}
	\label{fig:rebuttal_gt}
\end{figure}

\noindent{\textbf{Real-world samples we collected for testing.}} The dToF and image resolutions are $40\times30$ and $912\times684$. We use stereo matching methods to generate ground truth (Figure~\ref{fig:rebuttal_gt}), while the main and ultra-wide cameras on the mobile phone are used to form a stereo pair. We also manually filter failed samples and mask noisy regions. Lens distortion and baseline mismatch may slightly affect the epipolar geometry, leading to a global shift. While not perfect, we believe the metrics still offer a meaningful preliminary evaluation, as the SOTA stereo methods~\cite{igev++, fastacc, monster, foundationstereo} are more effective than common sensors, especially in complex regions targeted in our work.

\begin{figure}[htbp]
	\centering
	\includegraphics[width=\columnwidth]{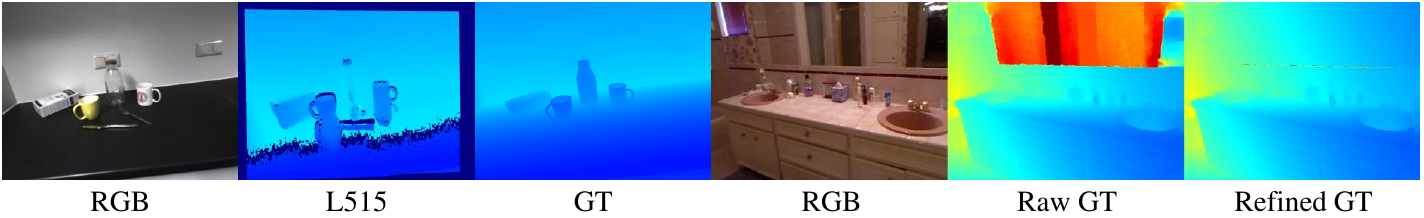}
	\caption{Examples of the Hammer (left) and Mirror3D-NYU (right) datasets. During testing, we sampled dToF points from L515 and Raw GT, respectively.}
	\label{fig:hammermirror}
\end{figure}

\noindent{\textbf{Hammer, Mirror3D and other real datasets for testing.}} To enhance representativeness, we simulate training data using a more common resolution—$30\times40$ for dToF and $480\times640$ for images. For evaluation, we simulate low-cost dToF data by sampling sparse depth from high-cost sensor measurements on various real-world datasets to assess whether our method can improve the sampled source. For Hammer~\cite{hammer} in Figure~\ref{fig:hammermirror}, we uniformly selected 721 frames from the total 7207 to form the test set. Sparse inputs are sampled from the raw RealSense L515 measurements and evaluated on the provided high-precision ground truth. For Mirror3D-NYU~\cite{mirror3d} in Figure~\ref{fig:hammermirror}, where the authors manually corrected depth values in specular regions of the NYUv2 dataset~\cite{nyuv2}, we sample sparse inputs from the raw ground truth and evaluate on the refined ground truth. For datasets without reliable ground truth in Figure~\ref{fig:realdataset}, such as NYUv2~\cite{nyuv2} and ARKitScenes~\cite{arkitscenes}, we sample sparse inputs from the ground truth and present qualitative results.

\noindent{\textbf{Evaluation Metrics}} We reported standard metrics including ${\delta}_{i}$, Rel, RMSE, $\text{log}_{10}$. To further evaluate performance at boundaries, we also reported edge-weighted mean absolute error (EWMAE)~\cite{mipi2023,ewmae}, which assigns greater weight to pixels with larger gradients when calculating MAE. The details are introduced in the supplementary material.

\subsection{Effectiveness of Our Training Strategy.}
\label{sec:exp_strategy}
We trained our model and a lightweight PENet (denoted as PENet*) using different simulation methods on Hypersim and evaluated on ZJU-L5 and our samples. For PENet*, we retain the original design but reduce the number of layers and channels to accelerate training. As a result, the parameters and FLOPs are reduced from 131M / 592G to 48M / 110G.
\begin{table}[ht]
	\centering
	\caption{\textbf{Performance on ZJU-L5 under different simulation methods.} The second row are reported by Deltar~\cite{deltar}.}
	\renewcommand{\arraystretch}{1}
	\setlength{\tabcolsep}{5pt}
	\resizebox{\linewidth}{!}{
		\begin{tabular}{lcccccc}
			\toprule
			Model  & Simulation & $\delta_1$ & $\delta_2$  & Rel & RMSE & $\log_{10}$\\
			\midrule
			CFPNet & Deltar & 0.883 & \underline{0.949} & 0.103 & \underline{0.431} & 0.047\\
			PENet  & Deltar & 0.807 & 0.914 & 0.161 & 0.498 & 0.065\\
			\midrule
			PENet* & Deltar & 0.815 & - & 0.152 & 0.510 & - \\
			PENet* & MIPI   & 0.865 & 0.929 & 0.118 & 0.493 & 0.061\\
			PENet* & Ours   & \underline{0.889} & \underline{0.949} & \underline{0.093} & 0.447 & \underline{0.046}\\
			\midrule
			Ours   & Deltar & 0.804 & 0.883 & 0.164 & 0.562 & 0.097 \\
			Ours   & MIPI   & 0.853 & 0.909 & 0.123 & 0.511 & 0.089 \\
			Ours   & Ours   & \textbf{0.933} & \textbf{0.972} & \textbf{0.075} & \textbf{0.350} & \textbf{0.034}\\
			\bottomrule
		\end{tabular}
	}
	\label{tab:comparison}
\end{table}

Table~\ref{tab:comparison} shows the quantitative results on ZJU-L5, our method significantly improves the performance of both models. Notably, PENet* outperforms the SOTA super-resolution method CFPNet on several metrics, demonstrating that our simulation strategy effectively narrows the gap between depth completion and super-resolution. Qualitative results on our samples in Figure~\ref{fig:implicit} also verified the effectiveness. We further analyze the impact of training datasets in Section~\ref{sec:exp_ablation}.

\begin{table}[ht]
	\centering
	\caption{\textbf{Performance on Hypersim under different training and testing simulations.} The results are based on ZJU-L5's dToF simulation, and our simulation is compatible with ideal inputs.}
	\renewcommand{\arraystretch}{1.3}
	\setlength{\tabcolsep}{3pt}
	\resizebox{\linewidth}{!}{%
		\begin{tabular}{c|cc|cc|cc|cc|cc|cc|cc}
			\toprule
			\multicolumn{1}{c|}{Training} 
			& \multicolumn{2}{c|}{Deltar} 
			& \multicolumn{2}{c|}{MIPI} 
			& \multicolumn{2}{c|}{Ours} 
			& \multicolumn{4}{c|}{Ours} 
			& \multicolumn{2}{c|}{Deltar} 
			& \multicolumn{2}{c}{MIPI} \\
			\midrule
			\multicolumn{1}{c|}{Testing} 
			& \multicolumn{2}{c|}{Deltar}  
			& \multicolumn{2}{c|}{MIPI} 
			& \multicolumn{2}{c|}{Ours} 
			& \multicolumn{2}{c|}{Deltar} 
			& \multicolumn{2}{c|}{MIPI} 
			& \multicolumn{4}{c}{Ours} \\
			\cmidrule(lr){1-15}
			Metrics & Rel & RMSE & Rel & RMSE & Rel & RMSE & Rel & RMSE & Rel & RMSE & Rel & RMSE & Rel & RMSE\\
			\midrule
			PENet* 
			& 0.050 & 0.402 
			& 0.060 & 0.434
			& 0.095 & 0.583 
			& 0.077 & 0.507 
			& 0.089 & 0.554 
			& 0.226 & 1.493 
			& 0.121 & 0.710 \\
			Ours  
			& \textbf{0.040} & \textbf{0.340}
			& \textbf{0.044} & \textbf{0.346} 
			& \textbf{0.069} & \textbf{0.436} 
			& \textbf{0.056} & \textbf{0.381} 
			& \textbf{0.065} & \textbf{0.419} 
			& 0.256	& 1.777 
			& 0.119& 0.723\\
			\bottomrule
		\end{tabular}
	}
	\label{tab:rebuttal_comparison}
\end{table}

We observed a more significant performance drop in our model when trained without our simulation method. To investigate this, we conducted experiments on Hypersim by evaluating the model under mismatched training and testing simulations. As shown in Table~\ref{tab:rebuttal_comparison}, we believe the drop reflects a typical generalization trade-off: models with stronger fitting capacity may overfit idealized training data (MIPI/Deltar), leading to a greater drop when tested on real-world data. We \textit{`reproduce'} the drop at the end of Table~\ref{tab:rebuttal_comparison}. A similar drop is also found between PENet* and the larger PENet. From the perspective of data fitting, our simulation lowers the risk of overfitting to the ideal distribution and improves real-world performance. In this case, stronger models still achieve better results with idealized inputs.

\subsection{Effectiveness of Our Depth Completion Model}
\label{sec:exp_model}
Since the anomaly detection method is not incorporated into the training process, we treat it as an independent part. In this section, we compare the depth completion model with other methods separately.

In addition to referencing results from published papers, we conducted additional experiments to ensure a fair and comprehensive comparison. The detailed test settings are as follows: \textbf{For monocular depth estimation (MDE)}, we evaluated two variants of Depth Anything v2: the large-metric version (DAv2-LM, fine-tuned on Hypersim) is directly tested, and the small-relative version (DAv2-SR, used in our method), where its inverse depth output was linearly fitted to the dToF signals. \textbf{For depth completion (DC)}, we performed two types of evaluations: (1) retraining existing methods with our strategy, including PENet*, the SOTA 2D method CFormer, and the 3D method BPNet on conventional benchmarks; (2) directly testing OMNI-DC, the SOTA generalizable method which trained across multiple modalities and varying sparsity levels. \textbf{For depth super-resolution (DS)}, we evaluated PromptDA that also integrate MDE models. Since it lacks training on absence, we linearly fitted the relative depth map from DAv2 to dToF and filled in the missing regions.

\noindent{\textbf{Results on ZJU-L5.}} Table~\ref{tab:zjul5_quantitative} presents the quantitative results on the ZJU-L5 dataset. The first seven rows are quoted from CFPNet, all trained on the NYUv2 dataset, while we supplemented the resting results. Generally, our method achieves substantial accuracy improvements for all metrics.

\begin{table}[ht]
	\centering
	\caption{\textbf{Quantitative comparison on ZJU-L5}. Different versions of our model achieved the \textbf{best} and \textit{second best} results. The best result among existing methods is \underline{underlined}. CFPNet is the published SOTA method focusing on this dataset.}
	\renewcommand{\arraystretch}{1.5} 
	\setlength{\tabcolsep}{4pt} 
	\resizebox{\linewidth}{!}{ 
		\begin{tabular}{lcc|ccccc}
			\toprule
			Method & Type & Pub & $\delta_1$ & $\delta_2$ & Rel & RMSE & $\log_{10}$ \\
			\midrule
			BTS\cite{bts} & MDE & arXiv19 & 0.739 & 0.914 & 0.174 & 0.523 & 0.079 \\
			AdaBins\cite{adabins} & MDE & CVPR21 & 0.770 & 0.926 & 0.160 & 0.494 & 0.073 \\
			PnP-Depth\cite{pnpdepth} & DS & ICRA19 & 0.805 & 0.904 & 0.144 & 0.560 & 0.068 \\
			PrDepth\cite{xia2020generating} & DS & CVPR20 & 0.800 & 0.926 & 0.151 & 0.460 & 0.063 \\
			PENet\cite{penet} & DC & ICRA21 & 0.807 & 0.914 & 0.161 & 0.498 & 0.065 \\
			Deltar\cite{deltar} & DS & ECCV22 & 0.853 & 0.941 & 0.123 & 0.436 & 0.051 \\
			CFPNet\cite{cfpnet} & DS & 3DV25 & 0.883 & \underline{0.949} & 0.103 & \underline{0.431} & 0.047 \\
			\midrule
			PENet*\cite{penet} & DC & ICRA21 & \underline{0.889} & \underline{0.949} & \underline{0.093} & 0.447 & \underline{0.046}\\
			CFormer\cite{completionformer} & DC & CVPR23 & 0.873 & 0.938 & 0.103 & 0.480 & 0.053 \\
			BPNet\cite{bpnet} & DC & CVPR24 & - & - & - & 0.671 & - \\
			DAv2 -LM\cite{depthanythingv2} & MDE & Neurips24 & 0.703 & 0.905 & 0.220 & 0.467 & 0.083 \\
			DAv2 -SR\cite{depthanythingv2} & MDE & Neurips24 & 0.869 & 0.937 & 0.109 & 0.480 & 0.063\\
			PromptDA\cite{promptda} & DS & CVPR25 & 0.885 & 0.947 & 0.096 & 0.444 & 0.051\\ 
			OMNI-DC\cite{omnidc} & DC & ICCV25 & 0.871 & 0.933 & 0.099 & 0.502 & 0.053 \\
			\midrule
			Ours-Small & DC & - & \textit{0.921} & \textit{0.963} & \textit{0.080} & \textit{0.379} & \textit{0.038} \\
			Ours-Large & DC & - & \textbf{0.933} & \textbf{0.972} & \textbf{0.075} & \textbf{0.350} & \textbf{0.034} \\
			\bottomrule
		\end{tabular}
	}
	\label{tab:zjul5_quantitative}
\end{table}

We found that due to changes in depth pattern and potential anomalies, many methods that are effective in traditional benchmarks are not well-suited for real-world dToF data. First, since dToF points are roughly uniformly distributed yet extremely sparse (only 0.02\% in ZJU-L5), 3D methods struggle to capture spatial interactions, and architectures sensitive to sparsity also suffer from performance degradation. Second, designs that assume the accuracy of sensors are sensitive to real-world noise, focusing solely on preserving and rapidly propagating the sparse measurement. Examples include the point embedding operation in the affinity propagation module and residual connections with the initial sparse depth map.

\begin{figure}[htbp]
	\centering
	\includegraphics[width=\columnwidth]{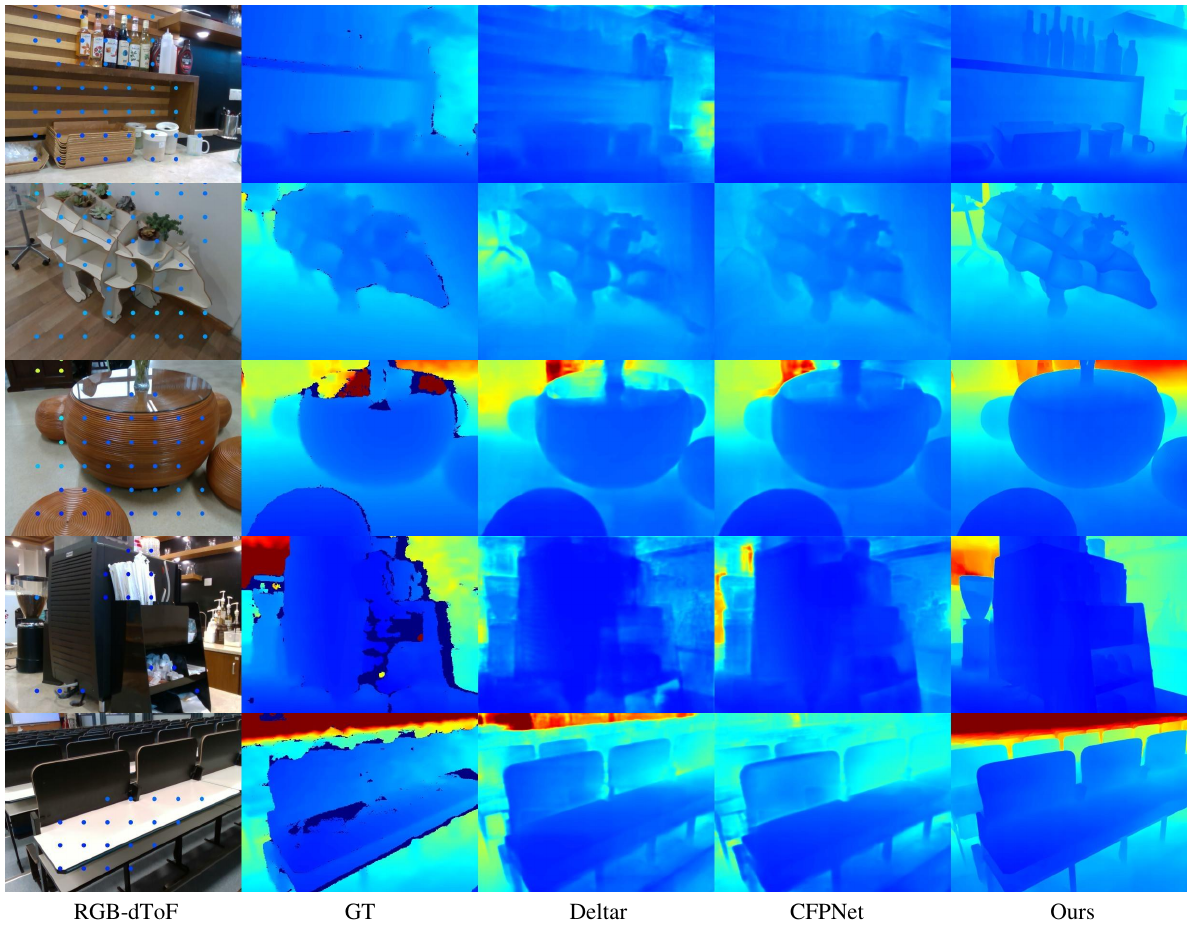}
	\caption{\textbf{Qualitative results on ZJU-L5}, our model further improves anomalies present in the ground truth}
	\label{fig:zju}
\end{figure}

In addition, as the limitations of real-world datasets mentioned before, we believe these metrics may not fully capture the performance of our method. As shown in Figure~\ref{fig:zju}, our model's predictions not only outperform existing methods but also further improve anomalies present in the ground truth, which leads to a decrease in metrics. More qualitative results are provided in the supplementary material.

\begin{figure}[htbp]
	\centering
	\includegraphics[width=\linewidth]{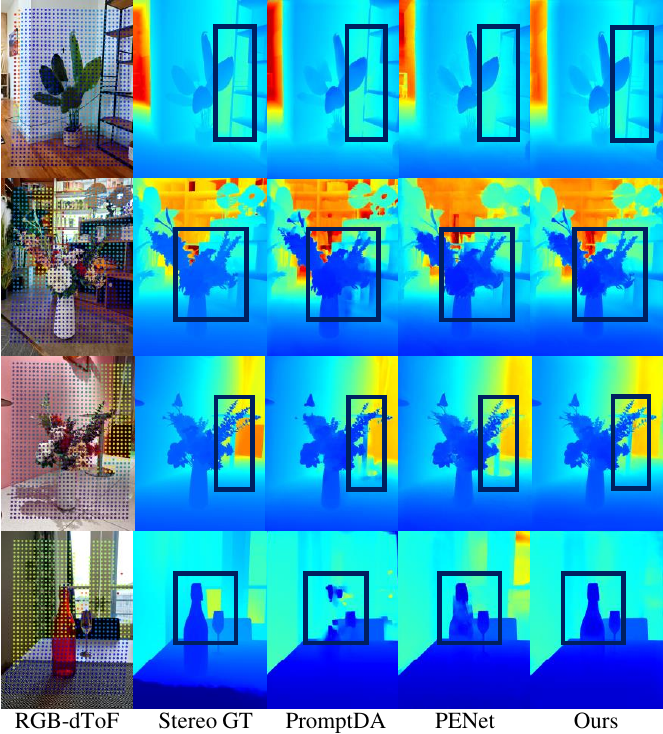}
	\caption{Qualitative results on our real-world samples.}
	\label{fig:compare}
\end{figure}

\noindent{\textbf{Results on our real-world samples.}} Due to the higher image resolution, we modified some methods to accelerate training. The ground truth is based on MonSter~\cite{monster}. As shown in Table~\ref{tab:sota}, our model achieves the best results. Figure~\ref{fig:compare} presents the qualitative comparison. Our method effectively integrates the MDE model, improving performance in detail and challenging regions. We observed that PENet achieves better edge prediction than PromptDA, which is consistent with the result of EWMAE in Table~\ref{tab:sota}. More qualitative results are provided in the supplementary material.

\begin{table}[ht]
	\centering
	\caption{\textbf{Quantitative comparison on our real-world samples.}}
	\renewcommand{\arraystretch}{1.2} 
	\setlength{\tabcolsep}{4pt} 
	\resizebox{\linewidth}{!}{%
		\begin{tabular}{lc|ccccc}
			\toprule
			Model & Pub &$\delta_1$ & $\delta_2$ & RMSE & Rel & EWMAE \\
			\midrule
			BPNet\cite{bpnet} & CVPR24 & - & - & 0.630 & - & - \\
			OMNI-DC\cite{omnidc} & ICCV25 & 0.593 & 0.768 & 0.643 & 0.292 & 0.195 \\
			DAv2 -SR\cite{depthanythingv2}& Neurips24 & 0.687 & 0.833 & 0.292 & 0.237 & 0.141 \\
			PENet*\cite{penet}& ICRA21 & 0.740 & 0.878 & 0.327 & 0.202 & \underline{0.139} \\ 
			CFormer*\cite{completionformer} & CVPR23 & 0.732 & 0.883 & 0.320 & 0.206 & 0.159 \\
			PromptDA\cite{promptda} & CVPR25 & \underline{0.761} & \underline{0.905} & \underline{0.268} & \underline{0.166} & 0.184 \\
			\midrule
			Ours-Small & - & \textit{0.785} & \textit{0.908} & \textit{0.251} & \textit{0.170} & \textit{0.117} \\
			Ours-Large & - & \textbf{0.790} & \textbf{0.911} & \textbf{0.226} & \textbf{0.155} & \textbf{0.108} \\
			\bottomrule
			\multicolumn{6}{l}{\footnotesize * indicates a lightweight version.}
		\end{tabular}
	}
	\label{tab:sota}
\end{table}

\noindent{\textbf{Results on Hammer.}} We directly sampled depth points from the raw outputs of L515 for testing, as shown in Table~\ref{tab:model_l515}. For semi-dense raw measurements, we only calculate the error among valid pixels, ignoring missing pixels. For depth completion models, we calculate the error for the entire image. The results further validate our previous conclusion: our training strategy substantially improves the performance of existing methods, while our proposed model achieves the best results. Notably, with only 1200 sampled points from L515, our model outperforms the original L515 measurements. This demonstrates that depth enhancement can not only bring low-cost sensors closer to the performance of high-precision devices, but also has the potential to surpass them.

\begin{table}[h]
	\centering
	\caption{\textbf{Quantitative comparison on Hammer,} testing dToF points are sampled from the raw L515 measurements.}
	\resizebox{\linewidth}{!}{%
		\begin{tabular}{c|cccccc}
			\toprule
			Sensor/Model & L515 & PENet* & PENet* & CFormer & Ours-S & Ours-L\\
			\midrule
			Train Simulation & - & Deltar & Ours & Ours & Ours & Ours\\
			\midrule
			$\delta_1$ & 0.963 & 0.946 & 0.980 & 0.981 & 0.981 & \textbf{0.982} \\
			Rel  & 0.036 & 0.060 & 0.044 & 0.042 & 0.038 & \textbf{0.034}\\
			RMSE & 0.060 & 0.093 & 0.061 & 0.058 & 0.048 & \textbf{0.044}\\
			\bottomrule
		\end{tabular}
	}
	\label{tab:model_l515}
\end{table}

\noindent{\textbf{Complexity Analysis.}}
Table~\ref{tab:complexity} presents the complexity of some methods, with FLOPs calculated at the resolution of $480\times640$. Our small version achieves the lowest computational cost and learnable parameters while still outperforming SOTA methods by leveraging a more lightweight network and computing relative depth maps at half resolution.

\begin{table}[htbp]
	\centering
	\caption{\textbf{Complexity comparison.} We separately list the depth completion and the MDE model (frozen) in our method.}
	\renewcommand{\arraystretch}{1.1} 
	\setlength{\tabcolsep}{5pt} 
	\resizebox{\linewidth}{!}{
		\begin{tabular}{cccccc}
			\toprule
			Method & Deltar & CFPNet & PENet* & DA-S & Ours-S \\
			\midrule
			Params \textit{(M)}& 18 & 20 & 48 & 24 & 6+\textit{24}  \\
			FLOPs \textit{(G)} & 42 & 46 & 110 & 47 & 26+\textit{13}  \\
			Time \textit{(ms)} & 44 & 57 & 24 & 18 & 24 \\
			\midrule
			Method & OMNI-DC &  CFormer & PromptDA & DA-L  & Ours-L \\
			\midrule
			Params \textit{(M)} & 84 & 81 & 337 & 335  & 12+\textit{24}\\
			FLOPs \textit{(G)}  & 398 & 380 & 713 & 674 & 64+\textit{47}\\
			Time \textit{(ms)}  & - & 86 & 120 & 116 & 36 \\
			\bottomrule
		\end{tabular}
	}
	\label{tab:complexity}
\end{table}

\begin{figure}[htbp]
	\centering
	\includegraphics[width=\linewidth]{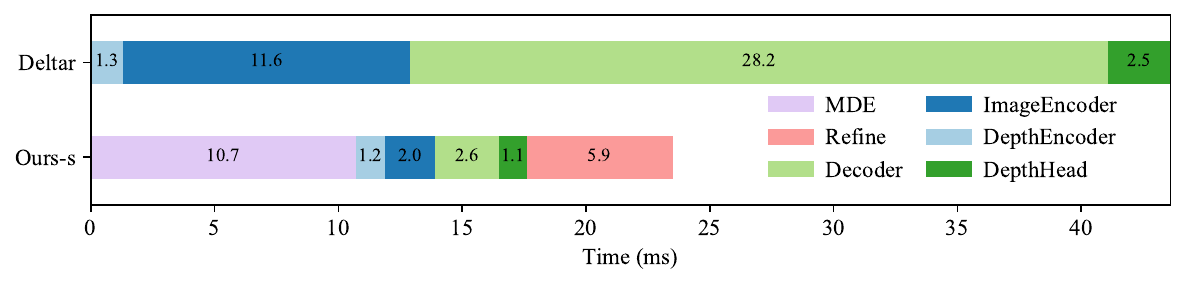}
	\caption{Runtime breakdown analysis of Deltar and Ours-S}
	\label{fig:submodule}
\end{figure}

We also analyze the inference speed on an NVIDIA 3090 GPU in Table~\ref{tab:complexity}. With larger FLOPs and parameters, our model is surprisingly faster than Deltar. A module-wise analysis in Figure~\ref{fig:submodule} reveals that the fusion module in Deltar’s decoder heavily relies on frequent tensor slicing operations, where irregular RGB and dToF patches are extracted based on coordinates for aggregation, which significantly limits the inference speed. By reformulating the task as completion rather than super-resolution, our method avoids the need for explicit position inputs and instead implicitly models spatial correspondences. This not only enhances robustness to inaccurate alignment but also offers benefits even when accurate correspondences are available, leading to improved efficiency.

\subsection{Effectiveness of Our Depth Anomaly Detection}
\label{sec:exp_filter}
Our anomaly detection method is mainly designed for the dToF sensors in our samples.
For the L5 sensor in the ZJU-L5 dataset, it is extremely low-cost to provide only $8\times8$ depth measurements, and exhibits anomalies primarily as signal loss rather than error value, detecting erroneous points becomes meaningless. Thus, we conducted experiments based on the setting of the Hammer dataset, and using the model in the last column of Table~\ref{tab:model_l515} as a baseline. 

Since our method focuses on sparse depth points, while most existing approaches are introduced to handle high-resolution dense depth maps, we design two types of experiments to ensure a fair comparison: \textbf{For Mirror3D~\cite{mirror3d} that predict anomaly masks from images:} We apply the generated masks to sampled sparse depth maps to remove considered erroneous points. The filtered depth maps are then fed into our depth completion model. \textbf{For end-to-end method TDCNet~\cite{tdcnet} that do not predict anomaly masks}: We construct two dense depth inputs: (1) the raw sensor measurements; and (2) the outputs of our depth completion model without applying our anomaly detection method.

\begin{table}[htbp]
	\centering
	\caption{\textbf{Quantitative comparison on Hammer of detection method.} Our method improves performance with minimal overhead.}
	\label{tab:filter_hammer}
	\resizebox{\linewidth}{!}{%
		\begin{tabular}{llcccccc}
			\toprule
			Input & Method & \makecell{Params \\ (M)}  & \makecell{FLOPs \\ (G)}  & \makecell{Time \\ (ms)} & $\delta_1$ & Rel & RMSE  \\
			\midrule
			\multirow{2}{*}{Sensor} 
			& -      & -    & -   & -   & 0.963 & 0.036 & 0.060 \\
			& TDCNet & 6.3  & 95  & 35  & 0.966 & 0.037 & 0.059 \\
			\midrule
			\multirow{4}{*}{\makecell{Our\\Model}}  
			& -        & 36   & 111 & 36  & 0.982 & 0.034 & 0.044 \\
			& TDCNet    & 6.3  & 95  & 35  & 0.984 & 0.034 & 0.043 \\
			& Mirror3D  & 14.5 & 29  & 91  & 0.981 & 0.034 & 0.046 \\
			& Ours      & +0   & +0  & +3  & \textbf{0.987} & \textbf{0.031} & \textbf{0.041} \\
			\bottomrule
		\end{tabular}
	}
\end{table}

Table~\ref{tab:filter_hammer} presents the results on the Hammer dataset. Since the dataset mainly consists of transparent objects and does not contain specular surfaces, Mirror3D did not lead to performance gains, and even caused degradation due to false positives. TDCNet improved performance when processing raw sensor inputs, but failed to handle anomalies that remain challenging for our model. Our method achieved further performance improvements with minor overhead, since our model already integrates the MDE model.

\begin{figure}[htbp]
	\centering
	\includegraphics[width=\linewidth]{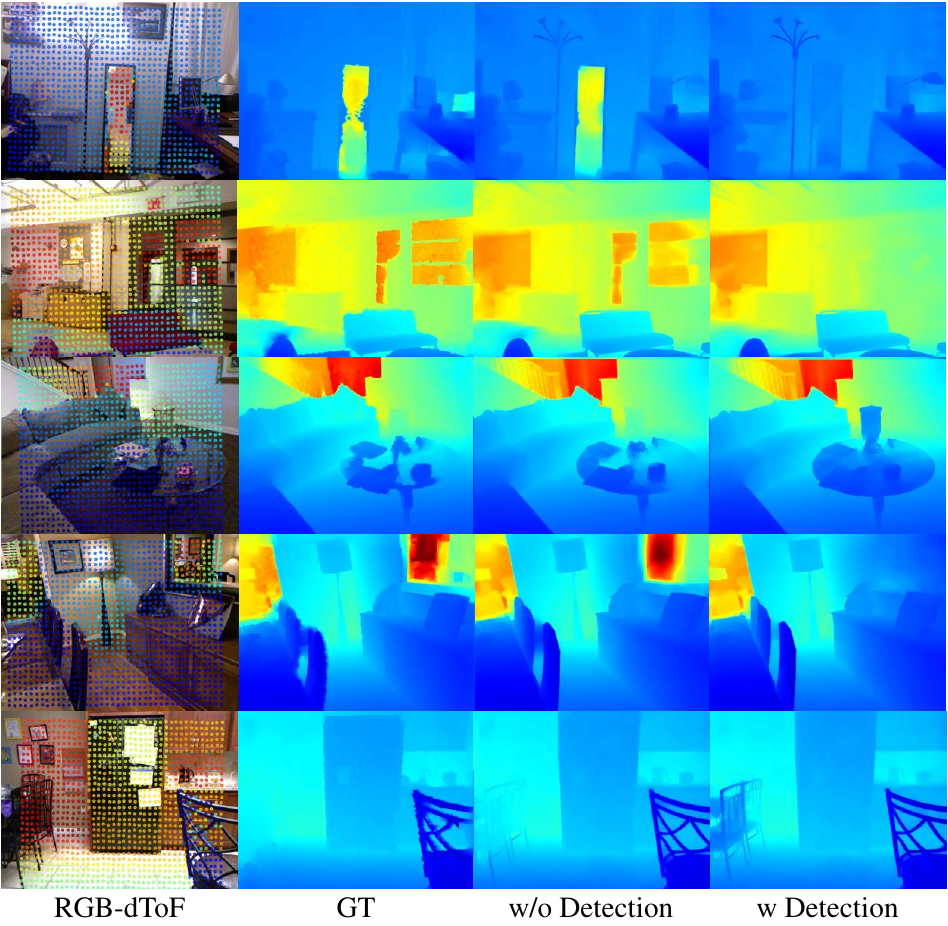}
	\caption{\textbf{Results of our model with anomaly detection on NYUv2,} dToF sampled from GT collected by Microsoft Kinect v1}
	\label{fig:filter_nyu}
\end{figure}

We further evaluated our method on the Mirror3D-NYU dataset. As shown in Table~\ref{tab:filter_mirror3d}, our method improves predictions on specular surfaces, outperforming Mirror3DNet. For the observed performance drop in other regions, we believe it can be attributed to our method detecting a broader range of anomalies, as the dataset only corrects depth values in mirrors; other types of errors remain unaddressed. This observation is further supported by the qualitative results in Figure~\ref{fig:filter_nyu}.

\begin{table}[htbp]
	\centering
	\caption{\textbf{Quantitative comparison on Mirror3D-NYU.} Our method achieves superior performance on mirror regions and detects broader anomalies beyond mirrors.}
	\label{tab:filter_mirror3d}
	\resizebox{\linewidth}{!}{%
		\begin{tabular}{l|ccc|ccc}
			\toprule
			\multirow{2}{*}{Method} & \multicolumn{3}{c|}{RMSE} & \multicolumn{3}{c}{Rel} \\
			\cmidrule(lr){2-4} \cmidrule(lr){5-7}
			& \textbf{Mirror} & Other & All & \textbf{Mirror} & Other & All \\
			\midrule
			Raw Signal         & 1.214 & 0.006 & 0.435 & 0.590 & 0.000 & 0.101 \\
			saic\cite{saic}		   & 1.081 & 0.074 & 0.391 & 0.556& 0.012 & 0.099 \\
			Mirror3DNet\cite{mirror3d}        & 0.891 & 0.077 & 0.309 & 0.454 & 0.008 & 0.074 \\
			Ours (w/o Detect)  & 1.053 & 0.127 & 0.439 & 0.520 & 0.024 & 0.113 \\
			Ours (w/ Detect)   & \textbf{0.601} & 0.168 & 0.327 & \textbf{0.260} & 0.030 & 0.079 \\
			\bottomrule
		\end{tabular}
	}
\end{table}

\begin{figure}[htbp]
	\centering
	\includegraphics[width=\linewidth]{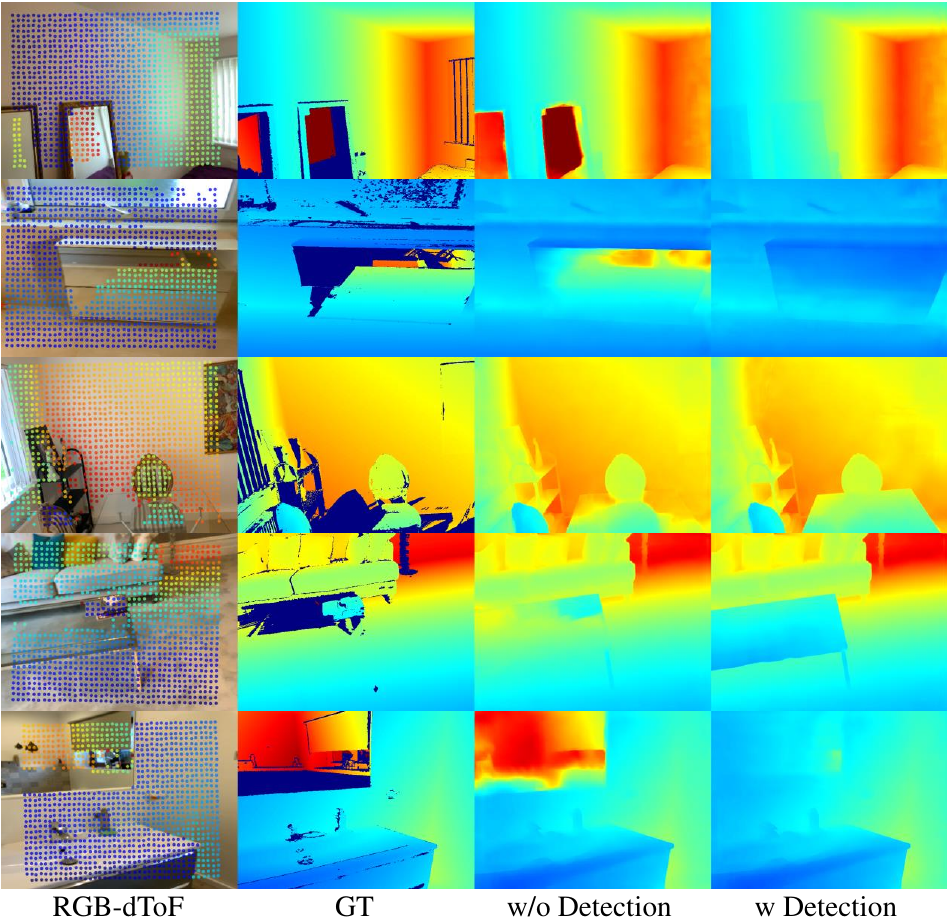}
	\caption{\textbf{Results of our model with anomaly detection on ARKitScenes,} dToF sampled from GT collected by Faro Focus S70.}
	\label{fig:filter_arkit}
\end{figure}

Figure~\ref{fig:filter_nyu} and Figure~\ref{fig:filter_arkit} show qualitative results on the NYUv2 and ARKitScenes. Although our model independently mitigates anomalies in localized regions, it struggles to handle large-area artifacts effectively. By incorporating the anomaly detection method, its predictions are significantly enhanced and, in some cases, can exceed the performance of high-precision sensors.

\subsection{Ablation Studies}
\label{sec:exp_ablation}
\noindent \textbf{Components of Simulation Method.} In Table~\ref{tab:ablationdegra}, we performed ablation studies on each component of our simulation method on the ZJU-L5 dataset. Since calibration errors are not considered, we validate it through qualitative results on real-world samples, provided in the supplementary material.

\begin{table}[htbp]
	\centering
	\caption{\textbf{Ablation studies about simulation method on ZJU-L5.} OD: Overall Distribution, SR: Specific Region, RA: Random Anomalies.}
	\resizebox{\linewidth}{!}{%
		\begin{tabular}{lccccc}
			\toprule
			Method & $\delta_1$ & $\delta_2$ & Rel  & RMSE  & $\log_{10}$ \\
			\midrule
			Standard & \textbf{0.933} & \textbf{0.972} & \textbf{0.075} & \textbf{0.350} & \textbf{0.034}\\
			w/o RA & 0.923 & 0.966 & 0.076 & 0.362 & 0.037\\
			w/o OD & 0.912 & 0.965 & 0.091 & 0.381 & 0.043\\
			w/o SR & 0.773 & 0.847 & 0.175 & 0.566 & 0.138\\
			w/o (RA + OD) & 0.905 & 0.965 & 0.102 & 0.395 & 0.045\\  
			w/o (OD + SR) & 0.780 & 0.855 & 0.191 & 0.641 & 0.168\\  
			\bottomrule
		\end{tabular}
	}

	\label{tab:ablationdegra}
\end{table}

\noindent \textbf{Depthor with Different MDE Models.} As shown in Table~\ref{tab:mde}, we replaced different MDE models in our model. We found that using more powerful MDE models does not significantly improve performance on ZJU-L5 compared to Hypersim, particularly in EWMAE, as the GT collected by RealSense D435 is blurred at the edges.

\begin{table}[htbp]
	\centering
	\caption{\textbf{Ablation studies on different MDE models}. The results on Hypersim are based on ZJU-L5's dToF simulation. R: relative; M: metric; S: small; B: base; L: large.}
	\resizebox{\linewidth}{!}{%
		\begin{tabular}{lcccccc}
			\toprule
			\multirow{3}{*}{Model} & \multicolumn{3}{c}{ZJU-L5} & \multicolumn{3}{c}{Hypersim} \\
			\cmidrule(lr){2-4} \cmidrule(lr){5-7}
			& RMSE & Rel & EWMAE & RMSE & Rel & EWMAE \\
			\midrule
			DAv2 -SR & 0.350 & 0.075 & 0.136 & 0.445 & 0.068 & 0.110 \\
			DAv2 -BR & 0.335 & 0.071 & 0.136 & 0.406 & 0.061 & 0.103 \\
			DAv2 -LR & 0.330 & 0.070 & 0.135 & 0.390 & 0.059 & 0.101 \\
			DAv2 -SM & 0.372 & 0.095 & 0.141 & 0.554 & 0.102 & 0.114 \\
			DAv1 -SR & 0.377 & 0.078 & - &-&-&- \\
			Midas-dpt-large & 0.414 & 0.086 & - &-&-&- \\
			\bottomrule
		\end{tabular}
	}
	\label{tab:mde}
\end{table}

\begin{figure}[htbp]
	\centering
	\includegraphics[width=\linewidth]{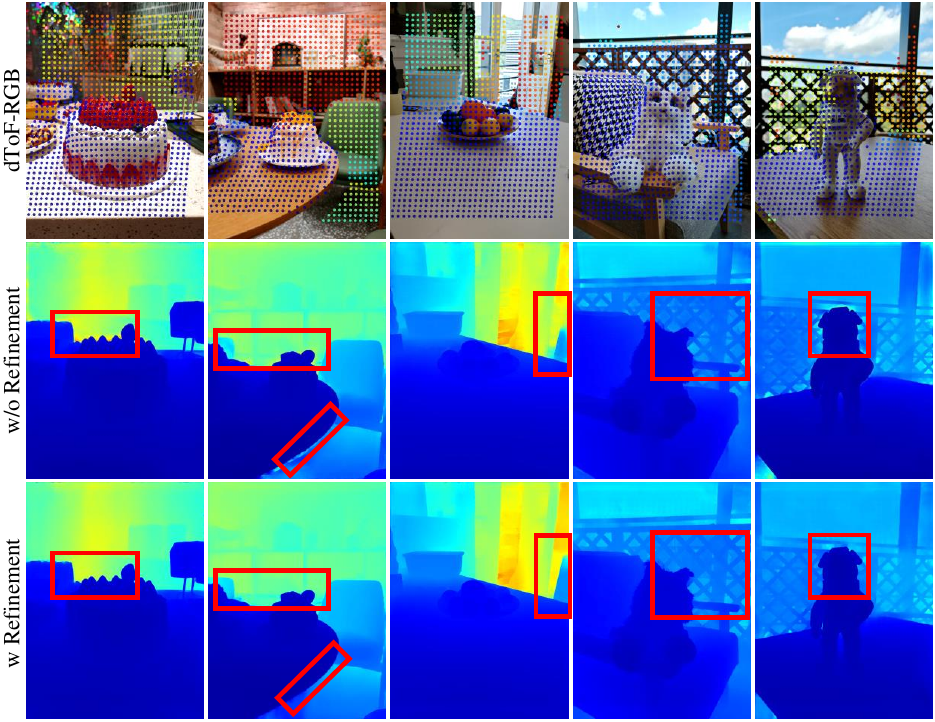}
	\caption{Refinement of mixed affinity propagation.}
	\label{fig:cspn}
\end{figure}

\noindent \textbf{Refinement of Mixed Affinity Propagation.} We analyze this module through quantitative metrics from synthetic datasets and qualitative results of real-world samples. As shown in Table~\ref{tab:cspn}, on the Hypersim dataset, refining the initial depth map in full resolution using affinity propagation improves the model’s overall performance, particularly on boundary-focused metric EWMAE. The qualitative results in Figure~\ref{fig:cspn} also reveal that this module effectively improves the model's performance in regions beyond the sensor’s FoV and at foreground-background boundaries, mitigating anomalies while enhancing prediction consistency.

\begin{table}[ht]
	\centering
	\caption{\textbf{Ablation studies about refinement.} The results on Hypersim are based on our real-world samples' dToF simulation}
	\resizebox{\linewidth}{!}{%
		\begin{tabular}{ccccccc}
			\toprule
			\multirow{2}{*}{Refine} & \multicolumn{2}{c}{Input Feature} & \multicolumn{3}{c}{Hypersim} & \multirow{2}{*}{Params.} \\
			\cmidrule(lr){2-3} \cmidrule(lr){4-6}
			& MDE & UNet & RMSE & REL & EWMAE & (\textit{M}) \\
			\midrule
			/ & / & / & 0.267 & 0.039 & 0.091 & - \\
			\checkmark & \checkmark & / & 0.269 & 0.039 & 0.087 & +0.048 \\
			\checkmark & / & \checkmark & 0.258 & 0.037 & 0.081 & +0.085 \\
			\checkmark & \checkmark & \checkmark & \textbf{0.248} & \textbf{0.034} & \textbf{0.079} & +0.122 \\
			\multicolumn{3}{c}{+ Point Embedding} & 0.328 & 0.038 & 0.098 & +0.122 \\
			\bottomrule
		\end{tabular}
	}
	\label{tab:cspn}
\end{table}

Furthermore, our experimental results demonstrate that due to the lack of scale information and the resolution differences, computing affinity solely based on ${F}_{mde}$ improves EWMAE but adversely affects scale metrics. However, the contextual information in ${F}_{mde}$ can still be leveraged to enhance ${F}_{unet}$. Additionally, the regional characteristics and anomalies of dToF signals conflict with the assumptions of point embedding, leading to performance degradation.

\begin{figure}[htbp]
	\centering
	\includegraphics[width=\linewidth]{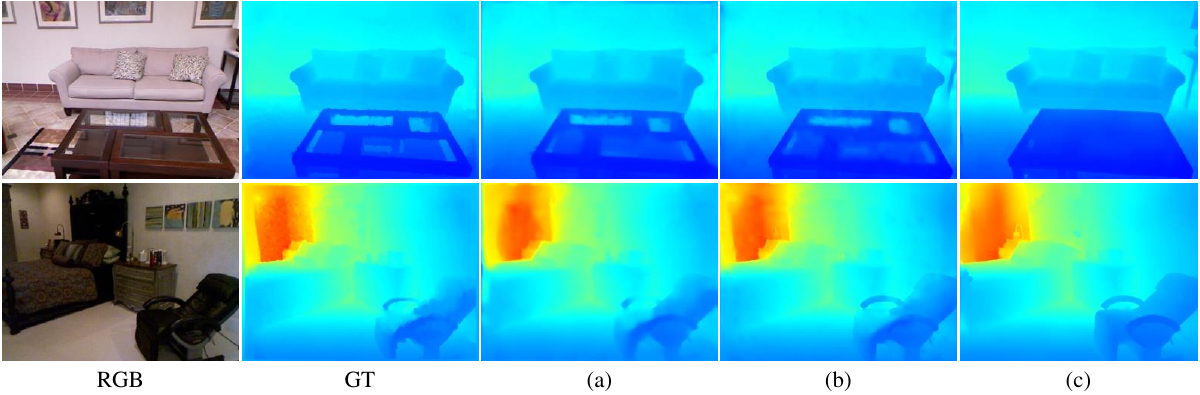}
	\caption{Prediction of our model under different training strategies on NYUv2.}
	\label{fig:complementarity}
\end{figure}

\noindent \textbf{Complementarity of Training Strategy and Model.} In Figure~\ref{fig:complementarity}, we present our model's predictions on the NYUv2 dataset under different training strategies: \textit{(a)} Trained on the NYUv2 dataset, model tends to disregard MDE outputs due to conflicts with the inaccurate ground truth; \textit{(b)} Trained on the Hypersim dataset without our simulation method, model extracts only contextual information from MDE, to propagate accurate depth points while neglecting global depth relationships; \textit{(c)} Our training strategy enhances performance through both global relationships and details.

\begin{figure}[htbp]
	\centering
	\includegraphics[width=\linewidth]{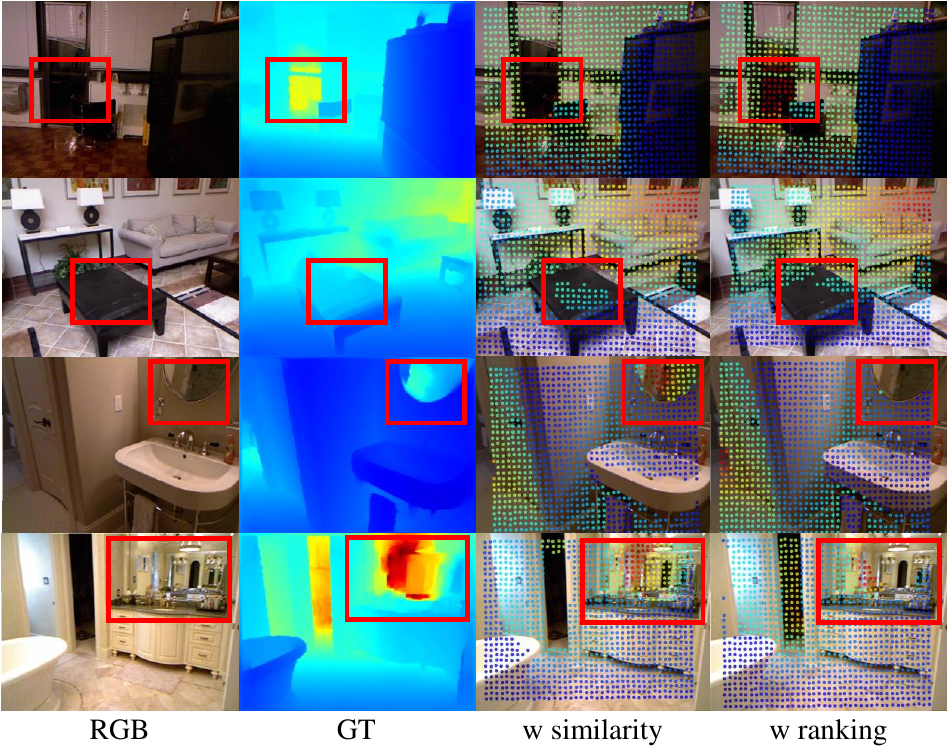}
	\caption{Detection results of different anomaly scores.}
	\label{fig:anomalyscore}
\end{figure}

\noindent \textbf{Component of Anomaly Score.} We separately use region similarity and ranking consistency for anomaly detection, and the results are shown in Figure~\ref{fig:anomalyscore}. The regional similarity focuses on local structural patterns, making it effective at detecting anomalies in high-frequency edge regions. However, it tends to fail in large, homogeneous areas such as mirrors, where depth values are highly consistent. In contrast, the ranking consistency captures global inconsistencies in depth ordering, allowing it to identify these anomalies. However, it may fail in certain cases as it ignores the absolute depth values, such as when already distant points become further away.

\noindent \textbf{Adaptive Thresholding to Reduce False Positives.} As shown in Table~\ref{tab:ablation_threshold}, when no noise is present (0\% in Hypersim), the original Otsu method produces a high false positive rate, leading to performance degradation. Using the Spearman rank correlation coefficient as a proxy for sensor reliability, our method enables adaptive threshold adjustment, which effectively suppresses false positives.

\begin{table}[ht]
	\centering
	\caption{Ablation Study about Threshold Function on the Hypersim dataset with 0\% error and on the real-world Hammer dataset.}
	\resizebox{\linewidth}{!}{
		\begin{tabular}{l|cccccc}
			\toprule
			\textbf{Datasets} & \multicolumn{3}{c}{Hypersim} & \multicolumn{3}{c}{Hammer}\\
			\midrule
			\textbf{Method} & ${\delta}_{1}$ & RMSE & REL & ${\delta}_{1}$ & RMSE & REL \\
			\midrule
			No Detection & 0.987 & 0.228 & 0.026 & 0.982 & 0.044 &0.034 \\
			Raw Otsu  & 0.975 & 0.306 & 0.035 & 0.987 & 0.046 & 0.036 \\
			Ours & 0.984 & 0.245 & 0.027 & 0.987 & 0.041 & 0.031 \\
			\bottomrule
		\end{tabular}
	}
	\label{tab:ablation_threshold}
\end{table}

We argue that the choice of anomaly threshold should be tailored to the model design and training strategy. When a model is effective at handling missing data or is trained on abundant signal loss samples, an aggressive threshold can be adopted to allow more false positives in exchange for improved overall accuracy. In contrast, when the model or training objective prioritizes robustness to error, a conservative threshold is preferable to avoid introducing false positives, since the model is already capable of small errors.

\noindent \textbf{Anomaly Detection for Other Depth Modality.} In addition to evaluating on dToF data with uniform distribution, we also tested on random distribution. On Hammer, we used OMNI-DC as the baseline and randomly sampled 500 sparse points from L515 raw measurements. As shown in Table~\ref{tab:enhance_hammer}, although OMNI-DC was trained with noisy points, our anomaly detection module still led to a substantial improvement.

\begin{table}[htbp]
	\centering
	\caption{Enhancing OMNI-DC with our detection method on the Hammer dataset.}
	\resizebox{\linewidth}{!}{%
		\begin{tabular}{c|c|cccc}
			\toprule
			Method/Sensor & Detection & ${\delta}_{1}$ & ${\delta}_{2}$ & RMSE & REL\\
			\midrule
			L515 & - & 0.963 & - & 0.060 & 0.036 \\
			\midrule
			\multirow{2}{*}{OMNI-DC}
			& / & 0.959 & 0.987 & 0.067 & 0.040\\
			& \checkmark & \textbf{0.971} & 0.995 & \textbf{0.057} & \textbf{0.033} \\
			\bottomrule
		\end{tabular}
	}
	\label{tab:enhance_hammer}
\end{table}

\noindent \textbf{Limitations.} The primary limitation of our depth completion model lies in its inference speed, which is not sufficient for real-time deployment on mobile devices. As suggested by our submodule analysis, potential improvements include quantizing the MDE model and optimizing the refinement module to further accelerate inference speed.

For the anomaly detection method, failure cases may occur in overly smooth scenes or when monocular depth estimation fails; representative examples are provided and discussed in the supplementary material. Additionally, since it needs to construct an $N \times N$ matrix for $N$ depth points, the computational cost may be prohibitive for high-resolution sensors such as LiDAR or those with dense measurements.

\section{Conclusion}
\label{sec:conclusion}
In this paper, we present DEPTHOR++, a comprehensive solution to real-world dToF enhancement that comprises both implicit robust learning and explicit anomaly detection. Unlike previous depth super-resolution methods, we reformulate the problem within depth completion to enable a more robust and flexible pipeline.

We introduce a noise-robust training strategy with a novel dToF simulation method on synthetic datasets, which addresses the performance bottlenecks and data scarcity related to real-world datasets in conventional settings. We also designed a novel network that effectively integrates MDE to enhance predictions in challenging regions. In addition, we propose a depth-point anomaly detection method that improves robustness by explicitly detecting and masking errors.

Our model with the proposed training strategy achieves state-of-the-art results on both the ZJU-L5 dataset and our real-world samples, with an improvement of 22\% and 11\%, respectively. On the Mirror3D-NYU dataset, our anomaly detection method further enhances model performance, surpassing the previous state-of-the-art by 37\%. On the Hammer dataset, using 1,200 sparse points sampled from RealSense L515, our full method surpasses the original L515 measurements with an average gain of 22\%. Qualitative results on diverse real-world datasets further demonstrate the effectiveness and generalizability of our method.

Extending our method to other depth sensors and exploring an end-to-end anomaly detection method for depth maps are promising directions for future research. 

\section*{Acknowledgments}
This research is supported by the National Key R\&D Program of China (2024YFE0217700), National Natural Science Foundation of China (62472184) and the Fundamental Research Funds for the Central Universities.

\bibliographystyle{IEEEtran}
\bibliography{DEPTHOR++}






 


\vspace{-40pt}


\vfill

\end{document}